\documentclass[10pt,twocolumn,letterpaper]{article}

\usepackage{cvpr}
\usepackage{times}
\usepackage{epsfig}
\usepackage{graphicx}
\usepackage{amsmath}
\usepackage{amssymb}
\usepackage{mathrsfs}
\usepackage{multirow}
\usepackage[table,xcdraw]{xcolor}
\usepackage{color}
\usepackage{subcaption}

\usepackage[breaklinks=true,bookmarks=false]{hyperref}

\cvprfinalcopy 

\def\cvprPaperID{****} 

\ifcvprfinal\fi
\begin{document}

\title{Geometry-Aware Symmetric Domain Adaptation for Monocular Depth Estimation}

\author{\mbox{Shanshan Zhao$^1$\quad\quad  Huan Fu$^1$\quad\quad Mingming Gong$^{2,3}$ \quad\quad Dacheng Tao$^1$}\\
$^1$UBTECH Sydney AI Center, School of Computer Science, FEIT, \\
University of Sydney, Darlington, NSW 2008, Australia\\
$^2$Department of Biomedical Informatics, University of Pittsburgh\\
$^3$Department of Philosophy, Carnegie Mellon University\\
{\tt\small \{szha4333@uni., hufu6371@uni., dacheng.tao@\}sydney.edu.au\quad\quad  mig73@pitt.edu}}

\maketitle

\begin{abstract}
{\it
Supervised depth estimation has achieved high accuracy due to the advanced deep network architectures. Since the groundtruth depth labels are hard to obtain, recent methods try to learn depth estimation networks in an unsupervised way by exploring unsupervised cues, which are effective but less reliable than true labels. An emerging way to resolve this dilemma is to transfer knowledge from synthetic images with ground truth depth via domain adaptation techniques. However, these approaches overlook specific geometric structure of the natural images in the target domain (i.e., real data), which is important for high-performing depth prediction. Motivated by the observation, we propose a geometry-aware symmetric domain adaptation framework (GASDA) to explore the labels in the synthetic data and epipolar geometry in the real data jointly. Moreover, by training two image style translators and depth estimators symmetrically in an end-to-end network, our model achieves better image style transfer and generates high-quality depth maps. The experimental results demonstrate the effectiveness of our proposed method and comparable performance against the state-of-the-art. Code will be publicly available at: \url{https://github.com/sshan-zhao/GASDA}.
}
\end{abstract}

\section{Introduction}
Monocular depth estimation~\cite{saxena2006learning,saxena2009make3d,eigen2014depth,ladicky2014pulling} has been an active research area in the field of computer vision.
Recent years have witnessed the great strides in this task, especially after deep convolutional neural networks (DCNNs) were exploited to estimate depth from a single image successfully~\cite{eigen2014depth}. Until now, there have been lots of follow-up works~\cite{liu2016learning,laina2016deeper,eigen2015predicting,li2015depth,xu2017multi,wang2015towards,fu2018deep} improving or extending this work. 
However, since the proposed deep models are trained in a fully supervised fashion, they require a large amount of data with ground truth depth, which is expensive to acquire in practice. To address this issue, unsupervised monocular depth estimation has been proposed~\cite{godard2017unsupervised,zhan2018unsupervised,garg2016unsupervised,xie2016deep3d}, using geometry-based cues and without the need of image-depth pairs during training. Unfortunately, this kind of method tends to be vulnerable to illumination change, occlusion and blurring and so on. 
\begin{figure}
\center
   \begin{subfigure}[b]{0.23\textwidth}
    \center
     \includegraphics[width=1\linewidth,height=0.5in]{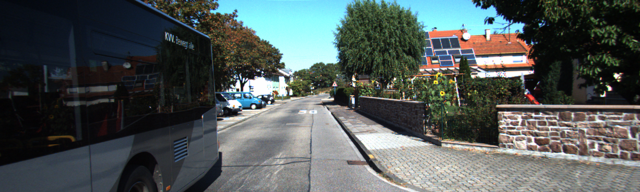}
     \setlength{\abovecaptionskip}{-10pt}
      \subcaption*{Real Image}
    \end{subfigure}%
    \hspace{0.1pt}
    \begin{subfigure}[b]{0.23\textwidth}
    \center
     \includegraphics[width=1\linewidth,height=0.5in]{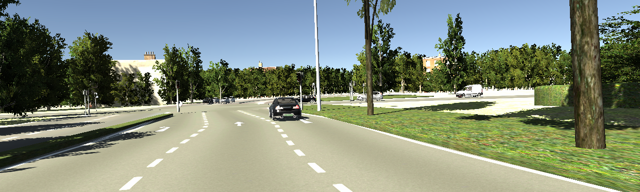}
     \setlength{\abovecaptionskip}{-10pt}
      \subcaption*{Synthetic Image}
    \end{subfigure}%

    \begin{subfigure}[b]{0.23\textwidth}
    \center
     \includegraphics[width=1.0\linewidth,height=0.5in]{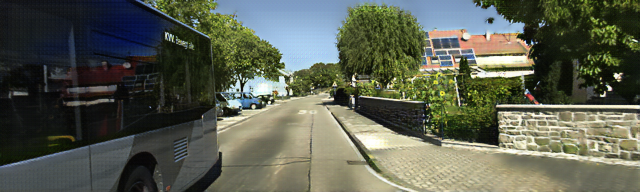}
     \setlength{\abovecaptionskip}{-10pt}
      \subcaption*{Real2Syn Image}
    \end{subfigure}
    \begin{subfigure}[b]{0.23\textwidth}
    \center
      \includegraphics[width=1.0\linewidth,height=0.5in]{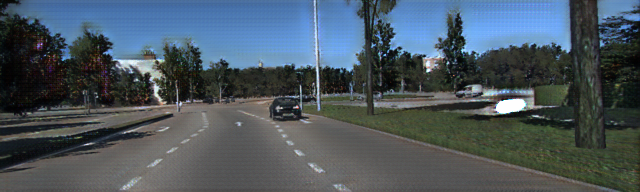}
      \setlength{\abovecaptionskip}{-10pt}
      \subcaption*{Syn2Real Image}
    \end{subfigure}

    \begin{subfigure}[b]{0.23\textwidth}
    \center
      \includegraphics[width=1.0\linewidth,height=0.5in]{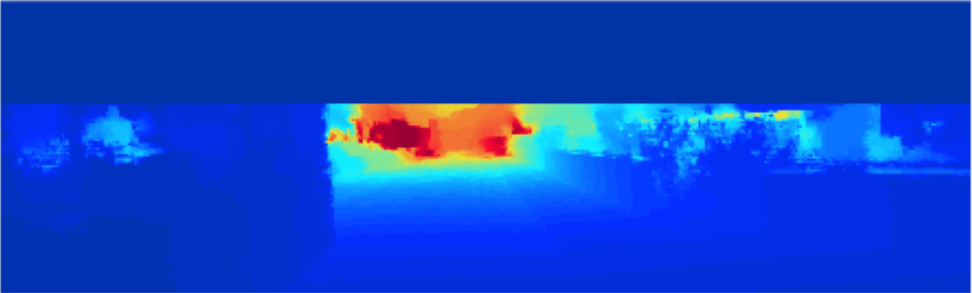}
      \setlength{\abovecaptionskip}{-10pt}
      \subcaption*{Ground Truth}
    \end{subfigure}
    \begin{subfigure}[b]{0.23\textwidth}
    \center
      \includegraphics[width=1.0\linewidth,height=0.5in]{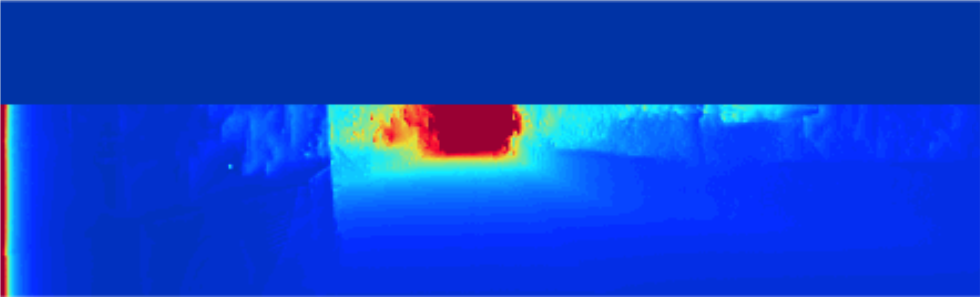}
      \setlength{\abovecaptionskip}{-10pt}
      \subcaption*{GASDA}
    \end{subfigure}
  \captionsetup{font={small}}
  \caption{Estimated Depth by GASDA. Top to bottom: input real image in the target domain (KITTI dataset~\cite{menze2015object}) and synthetic image for training (vKITTI dataset~\cite{gaidon2016virtual}), intermediate generated images in our approach, ground truth depth map and estimated depth map using proposed GASDA.}
  \label{fig:show} 
\end{figure}
Compared to real-world data, synthetic data is much easier to obtain the depth map. As a result, some works propose to exploit synthetic data for visual tasks~\cite{lai2017semi,long2013transfer,dosovitskiy2015flownet}. However, due to domain shift from synthetic to real, the model trained on synthetic data often fails to perform well on real data. To deal with this issue, domain adaptation techniques are utilized to reduce the discrepancy between datasets/domains~\footnote{We will use {\it domain} and {\it dataset} interchangeably
for the same meaning in most cases.}~\cite{atapour2018real,chen2018domain,long2013transfer}.

Existing works~\cite{atapour2018real,kundu2018adadepth,zheng2018t2net} using synthetic data via domain adaptation have achieved impressive performance for monocular depth estimation. These approaches typically perform domain adaptation either based on synthetic-to-realistic translation or inversely.
However, due to the lack of paired images, the image translation function usually introduces undesirable distortions in addition to the style change. The distorted image structures significantly degrade the performance of successive depth prediction. Fortunately, the unsupervised cues in the real images, for example, stereo pairs, produces additional constraints on the possible depth predictions. Therefore, it is essential to simultaneously explore both synthetic and real images and the corresponding depth cues for generating higher-quality depth maps.

Motivated by the above analysis, we propose a {\bf Geometry-Aware Symmetric Domain Adaptation Network (GASDA)} for unsupervised monocular depth estimation. This framework consists of two main parts, namely symmetric style translation and monocular depth estimation. Inspired by CycleGAN~\cite{CycleGAN2017}, our GASDA employs both synthetic-to-realistic and realistic-to-synthetic translations coupled with a geometry consistency loss based on the epipolar geomery of the real stereo images.
Our network is learned by groundtruth labels from the synthetic domain as well as the epipolar geometry of the real domain. Additionally, the learning process in the real and synthetic domains can be regularized by enforcing consistency on the depth predictions. By training the style translation and depth prediction networks in an end-to-end fashion, our model is able to translate  images without distorting the geometric and semantic content, and thus achieves better depth prediction performance. Our contributions can be summarized as follows:

\begin{itemize}
\item We propose an end-to-end  domain adaptation framework for monocular depth estimation. The model can generate high-quality results for both image style translation and depth estimation.
\item We show that training the monocular depth estimator using ground truth depth in the synthetic domain coupled with the epipolar geometry in the real domain can boost the performance.
\item We demonstrate the effectiveness of our method on KITTI dataset~\cite{menze2015object} and the generalization performance on Make3D dataset~\cite{saxena2009make3d}.

\end{itemize}
\begin{figure*}
\centering
    \begin{subfigure}[b]{0.175\textwidth}
      \centering
      \includegraphics[width=1.0in,height=0.62in]{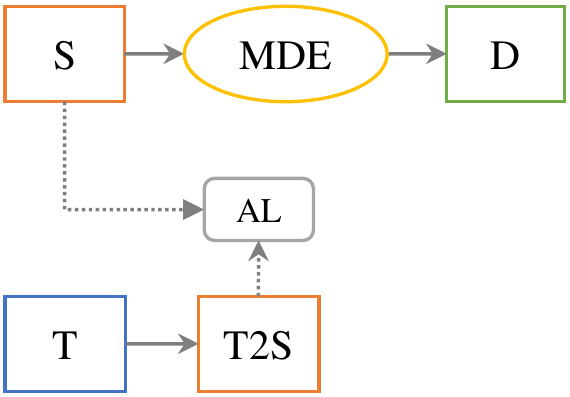}
      \subcaption*{Basic}
    \end{subfigure}%
    \hspace{-1pt}
    \vline
    \begin{subfigure}[b]{0.224\textwidth}
      \centering
      \includegraphics[width=1.3in, height=0.62in]{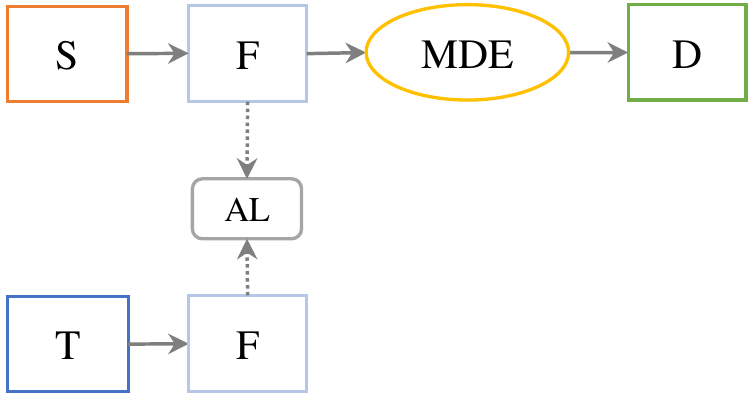}
      \subcaption*{AdaDepth~\cite{kundu2018adadepth}}
    \end{subfigure}
     \hspace{-5pt}
    \vline
    \hspace{-5pt}
    \begin{subfigure}[b]{0.28\textwidth}
      \centering
      \includegraphics[width=1.6in,height=0.62in]{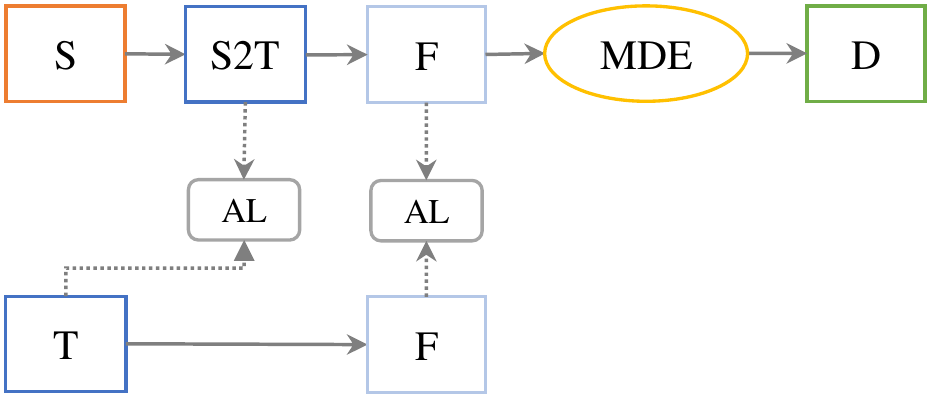}
      \subcaption*{T2Net~\cite{zheng2018t2net}}
    \end{subfigure}
    \hspace{-8pt}
    \vline
    \hspace{-7pt}
    \begin{subfigure}[b]{0.3\textwidth}
      \centering
      \includegraphics[width=1.7in,height=0.62in]{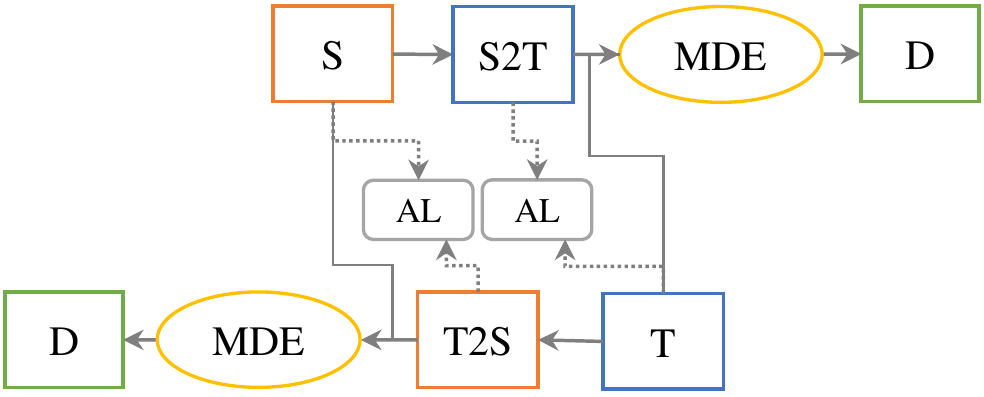}
      \subcaption*{GASDA}
    \end{subfigure}
  \captionsetup{font={small}}
  \caption{Different frameworks for monocular depth estimation using domain adaptation. Left to right: approach proposed in ~\cite{kundu2018adadepth},~\cite{zheng2018t2net} and this work respectively. S, T, F, S2T (T2S) and D represent the synthetic data, real data, extracted feature, generated data, and estimated depth. AL and MDE mean adversarial loss and monocular depth estimation, respectively. Compared with existing methods, our approach utilizes real stereo data and takes into account synthetic-to-real as well as real-to-synthetic during translation.}
  \label{fig:different_framework} 
\end{figure*}
\section{Related Work}

{\bf Monocular Depth Estimation} has been intensively studied over the past decade due to its crucial role in 3D scene understanding. Typical approaches sought the solution by exploiting probabilistic graphical models ({\it e.g.}, MRFs) ~\cite{saxena2009make3d,saxena2006learning,liu2010single}, and non-parametric techniques~\cite{liu2014discrete,karsch2014depth,liu2011sift}. However, these methods showed some limitations in performance and efficiency because of the employment of hand-crafted features and the low inference speed.

Recent studies demonstrated that high-performing depth estimators can be obtained relying on deep convolutional neural networks (DCNNs)~\cite{eigen2014depth,liu2016learning,he2018learning,xu2018structured,repala2018dual,qi2018geonet,cao2016estimating,laina2016deeper,roy2016monocular,chen2016single}. Eigen {\it et al.}~\cite{eigen2014depth} developed the first end-to-end deep model for depth estimation, which consists of a coarse-scale network and a fine-scale network. To exploit the relationships among image features, Liu {\it et al.}~\cite{liu2016learning} proposed to integrate continuous CRFs with DCNNs at super-pixel level. While previous works considered depth estimation as a regression task, Fu {\it et al.}~\cite{fu2018deep} solved depth estimation in the discrete paradigm by proposing an ordinal regression loss to encourage the ordinal competition among depth values.

A weakness of supervised depth estimation is the heavy requirement of annotated training images. To mitigate the issue, several notable attempts have investigated depth estimation in an unsupervised manner by means of stereo correspondence. Xie {\it et al.}~\cite{xie2016deep3d} proposed the Deep3D network for 2D-to-3D conversion by minimizing the pixel-wise reconstruction error. This work motivated the development of subsequent unsupervised depth estimation networks~\cite{garg2016unsupervised,godard2017unsupervised,yin2018geonet,zhou2017unsupervised}. In specific, Garg {\it et al.}~\cite{garg2016unsupervised} showed that unsupervised depth estimation could be recast as an image reconstruction problem according to the epipolar geometry. Following Garg {\it et al.}~\cite{garg2016unsupervised}, several later works improved the structure by exploiting left-right consistency~\cite{godard2017unsupervised}, learning depth in a semi-supervised way~\cite{kuznietsov2017semi}, and introducing temporal photometric constraints~\cite{zhan2018unsupervised}.

{\bf Domain Adaptation}~\cite{pan2010survey} aims to address the problem that the model trained on one dataset fails to generalize to another due to {\it dataset bias}~\cite{torralba2011unbiased}. In this community, previous works either learn the domain-invariant representations on a feature space ~\cite{ganin2015unsupervised,ganin2016domain,long2013transfer,ajakan2014domain,gong2016domain,gong2018causal,li2018deep}  or learn a mapping between the source and target domains at feature or pixel level~\cite{saenko2010adapting,sun2016deep,gong2012geodesic,zhang2013domain}. For example, Long {\it et al.}~\cite{long2013transfer} aligned feature distribution across the source and target domains by minimizing a Maximum Mean Discrepancy (MMD)~\cite{gretton2012kernel}. Tzeng {\it et al.}~\cite{tzeng2014deep} proposed to minimize MMD and the classification error jointly in a DCNN framework. Sun {\it et al.}~\cite{sun2016deep} proposed to match the mean and covariance of the two domain's deep features using the Correlation Alignment (CORAL) loss~\cite{sun2016return}.

Coming to domain adaptation for depth estimation, Atapour {\it et al.}~\cite{atapour2018real} developed a two-stage framework.  In specific, they first learned a translator to stylize the natural images so as to make them indistinguishable with the synthetic images, and then trained a depth estimation network using the original synthetic images in a supervised manner. Kundu {\it et al.}~\cite{kundu2018adadepth} proposed a content congruent regularization method to tackle the model collapse issue caused by domain adaptation in high dimensional feature space. Recently, Zheng {\it et al.}~\cite{zheng2018t2net} developed an end-to-end adaptation network, {\it i.e.} ${\rm T^2Net}$,  where the translation network and the depth estimation network are optimized jointly so that they can improve each other. However, these works overlooked  the geometric structure of the natural images from the target domain, which has been demonstrated significant for depth estimation~\cite{godard2017unsupervised,garg2016unsupervised}. Motivated by the observation, we propose a novel geometry-aware symmetric domain adaptation network, {\it i.e.}, GASDA, by exploiting the epipolar geometry of the stereo images.
The differences  between GASDA and previous depth adaptation approaches~\cite{kundu2018adadepth,zheng2018t2net} are shown in Figure~\ref{fig:different_framework}.
\section{Method}
\label{method}
\subsection{Method Overview}
Given a set of $N$ synthetic image-depth pairs $\{(x_s^i, y_s^i)\}_{i=1}^N$ ({\it i.e.}, source domain $X_s$), our goal here is to learn a monocular depth estimation model which can accurately predict depth for natural images contained in $X_t$ ({\it i.e.}, target domain). It is difficult to guarantee the model generalize well to the real data~\cite{atapour2018real,zheng2018t2net} due to the domain shift. We thus provide a remedy by exploiting the epipolar geometry between stereo images and developing a geometry-aware symmetric domain adaptation network (GASDA). Our GASDA consists of two main parts like existing works, including the style transfer network and the monocular depth estimation network. 

\begin{figure*}
\begin{center}
 \includegraphics[width=0.75\linewidth,height=2.96in]{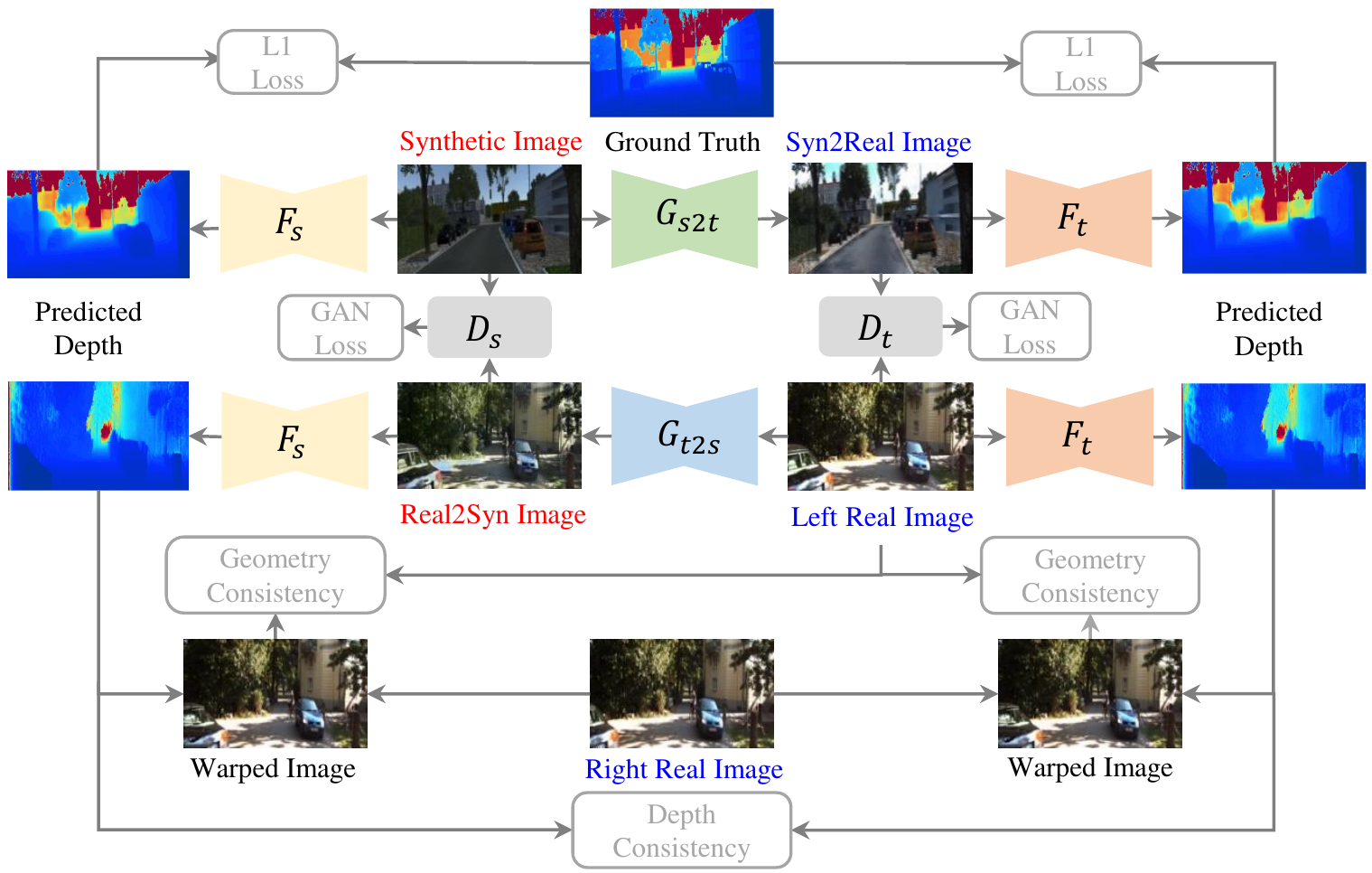}
\end{center}
\setlength{\abovecaptionskip}{-1.0pt}
\captionsetup{font={small}}
   \caption{The proposed framework in this paper.  It consists of two main parts: image style translation and monocular depth estimation. i) Style translation network, incorporating two generators ({\it i.e.}, $G_{s2t}$ and $G_{t2s}$) and two discriminators ({\it i.e.}, $D_t$ and $D_s$), is based on CycleGAN~\cite{CycleGAN2017}. ii) Monocular depth estimation network contains two complementary sub-networks ({\it i.e.}, $F_s$ and $F_t$). We omit the side outputs, for brevity. More details can be found in Section~\ref{method}, Section~\ref{sec:ex}.}
\label{fig:framework}
\end{figure*}

Specifically, unlike~\cite{atapour2018real,zheng2018t2net,kundu2018adadepth}, we consider both synthetic-to-real~\cite{zheng2018t2net} and real-to-synthetic translations~\cite{atapour2018real,kundu2018adadepth}. As a result, we can train two depth estimators $F_s$ and $F_t$ on the original synthetic data ($X_s$) and the generated realistic data ($G_{s2t}(X_s)$) using the generator $G_{s2t}$ in supervised manners, respectively. These two models are complementary, since $F_s$  has clean training set $X_s$ but dirty test set $G_{t2s}(X_t)$ generated by the generator $G_{t2s}$ with noises, such as distortion and blurs, caused by unsatisfied translation, and vise verse for $F_t$.
Nevertheless, because the depth information is rather relevant to specific scene geometry which might be different between source and target domains, the models trained on $X_s$ or $G_{s2t}(X_s)$ still could fail to perform well on $G_{t2s}(X_t)$ or $X_t$. To provide a solution, we exploit the epipolar geometry of real  stereo pairs $\{(x_{t_l}^i, x_{t_r}^i)\}_{i=1}^M$ ($x_{t_l}^i$ and $x_{t_r}^i$ represent the left and right image respectively\footnote{We will omit the subscript $l$ of $t_l$ for the left image in most cases.}) during training to encourage $F_t$ and $F_s$ to capture the relevant geometric structure of target/real data. In addition, we introduce an additional depth consistency loss to enforce the predictions from $F_t$ and $F_s$ are consistent in local regions. The overall framework of GASDA is illustrated in Figure~\ref{fig:framework}. For simplicity, we will omit the superscript $i$ in most cases.
\subsection{GASDA}
\noindent {\bf Bidirectional Style Transfer Loss} Our goal here is to learn the bidirectional translators $G_{s2t}$ and $G_{t2s}$ to bridge the gap between the source domain (synthetic) $X_s$ and the target domain (real) $X_t$. Specifically, taking $G_{s2t}$ as an example, we expect the $G_{s2t}(x_s)$ to be indistinguishable from real images in $X_t$. We thus employ a discriminator $D_t$, and train $G_{s2t}$ and $D_t$ in an adversarial fashion by performing a minimax game following~\cite{goodfellow2014generative}. The adversarial losses are expressed as:
\begin{equation}
\begin{aligned}
\mathcal{L}_{gan}(G_{s2t}, D_t, X_t, X_s)=&\mathbb{E}_{x_t\thicksim X_t}[D_t(x_t)-1]+\\
&\mathbb{E}_{x_s\thicksim X_s}[D_t(G_{s2t}(x_s))],\\
\mathcal{L}_{gan}(G_{t2s}, D_s, X_t, X_s)=&\mathbb{E}_{x_s\thicksim X_s}[D_s(x_s)-1]+\\
&\mathbb{E}_{x_t\thicksim X_t}[D_s(G_{t2s}(x_t))].
\end{aligned}
\end{equation}

Unluckily, the vanilla GANs suffer from mode collapse. To provide a remedy and ensure the input images and the output images paired up in a meaningful way, we utilize the cycle-consistency loss~\cite{CycleGAN2017}. Specifically, when feeding an image $x_s$ to $G_{s2t}$ and $G_{t2s}$ orderly, the output should be a reconstruction of $x_s$, and vice versa for $x_t$, {\it i.e.} $G_{t2s}(G_{s2t}(x_s))\thickapprox x_s$ and $G_{s2t}(G_{t2s}(x_t))\thickapprox x_t$. The cycle consistency loss has the form as:
\begin{equation}
\begin{aligned}
\mathcal{L}_{cyc}(G_{t2s}, G_{s2t})&=\mathbb{E}_{x_s\thicksim X_s}[||G_{t2s}(G_{s2t}(x_s))-x_s||_1]\\
&+\mathbb{E}_{x_t\thicksim X_t}[||G_{s2t}(G_{t2s}(x_t))-x_t||_1].
\end{aligned}
\end{equation}

Apart from the adversarial loss and cycle consistency loss, we also employ an identity mapping loss~\cite{taigman2016unsupervised} to encourage the generators to preserve geometric content. The identity mapping loss is given by:
\begin{equation}
\begin{aligned}
\mathcal{L}_{idt}(G_{t2s}, G_{s2t}, X_s, X_t)&=\mathbb{E}_{x_s\thicksim X_s}[||G_{t2s}(x_s)-x_s||_1]\\
&+\mathbb{E}_{x_t\thicksim X_t}[||G_{s2t}(x_t)-x_t||_1].
\end{aligned}
\end{equation}

The full objective for the bidirectional style transfer is as follow:
\begin{equation}
\begin{aligned}
\mathcal{L}_{trans}(G_{t2s}, G_{s2t}, D_t, D_s)&=\mathcal{L}_{gan}(G_{s2t}, D_t, X_t, X_s)\\
&+\mathcal{L}_{gan}(G_{t2s}, D_s, X_t, X_s)\\
&+\lambda_1 \mathcal{L}_{cyc}(G_{t2s}, G_{s2t})\\
&+\lambda_2 \mathcal{L}_{idt}(G_{t2s}, G_{s2t}, X_t, X_s)
\end{aligned}
\end{equation}
where $\lambda_1$ and $\lambda_2$ are the trade-off parameters.

\noindent {\bf Depth Estimation Loss} We can now render the synthetic images to the ``style" of the target domain (KITTI), and then capture a new dataset $X_{s2t} = G_{s2t}(X_s)$. We train a depth estimation network $F_t$ on $X_{s2t}$ in a supervised manner using the provided ground truth depth in the synthetic domain $X_s$. Here, we minimize the $\ell_1$ distance between the predicted depth $\tilde{y}_{ts}$ and ground truth depth $y_s$:
\begin{equation}
\begin{aligned}
&\mathcal{L}_{tde}(F_t, G_{s2t})=||y_s-\tilde{y}_{ts}||.
\end{aligned}
\end{equation}

In addition to $F_t$, we also train a complementary depth estimator $F_s$ on $X_s$ directly with the $\ell_1$ loss:
\begin{equation}
\begin{aligned}
&\mathcal{L}_{sde}(F_s)=||y_s-\tilde{y}_{ss}||
\end{aligned}
\end{equation}
where $\tilde{y}_{ss}=F_s(x_s)$ is the output of $F_s$. Both the $F_s$ and $F_t$ are important backbones to alleviate the issue of geometry and semantic inconsistency coupled with the subsequent losses.
The full depth estimation loss is expressed as:
\begin{equation}
\begin{aligned}
&\mathcal{L}_{de}(F_t, F_s, G_{s2t})=\mathcal{L}_{sde}(F_s)+\mathcal{L}_{tde}(F_t, G_{s2t}).
\end{aligned}
\end{equation}
\noindent {\bf Geometry Consistency Loss} Combining the components above, we have already formulated a naive depth adversarial adaptation framework. However, the $G_{s2t}$ and $G_{t2s}$ are usually imperfect, which would make the predictions $\tilde{y}_{st}=F_s(G_{t2s}(x_t))$ and $\tilde{y}_{tt}=F_t(x_t)$ unsatisfied. Besides, previous depth adaptation approaches overlook the specific physical geometric structure which may vary from scenes/datasets. Our main objective is to accurately estimate depth for real scenes, so we consider the geometric structure of the target data in the training phase. To this end, we present a geometric constraint on $F_t$ and $F_s$ by exploiting the epipolar geometry of real stereo images and unsupervised cues. Specifically, we generate an inverse warped image from the right image using the predicted depth, to reconstruct the left. We thus combine an $\ell_1$ with single scale SSIM~\cite{wang2004image} term as the geometry consistency loss to align the stereo images:
\begin{equation}
\label{eq:gc}
\begin{aligned}
&\mathcal{L}_{tgc}(F_t)=\eta\frac{1-SSIM(x_{t}, x^{'}_{tt})}{2}+\mu||x_{t}-x^{'}_{tt}||, \\
&\mathcal{L}_{sgc}(F_s, G_{t2s})=\eta\frac{1-SSIM(x_{t}, x^{'}_{st})}{2}+\mu||x_{t}-x_{st}^{'}||,\\
&\mathcal{L}_{gc}(F_t, F_s, G_{t2s})=\mathcal{L}_{tgc}(F_t)+\mathcal{L}_{sgc}(F_s, G_{t2s})
\end{aligned}
\end{equation}
where $\mathcal{L}_{gc}$ represents the full geometry consistency loss, $\mathcal{L}_{tgc}$ and $\mathcal{L}_{sgc}$ denote the geometry consistency loss of $F_t$ and $F_s$ respectively. ${x}^{'}_{tt}$ (${x}^{'}_{st}$) is the inverse warp of $x_{t_r}$ using bilinear sampling~\cite{jaderberg2015spatial} based on the estimated depth map $y_{tt}$ ($y_{st}$), the baseline distance between the cameras and the camera focal length~\cite{godard2017unsupervised}. In our experiments, $\eta$ is set to be $0.85$, and $\mu$ is $0.15$.\\
\noindent {\bf Depth Smoothness Loss} To encourage depths to be consistent in local homogeneous regions, we exploit an edge-aware depth smoothness loss:
\begin{equation}
\begin{aligned}
\mathcal{L}_{ds}(F_t, F_s, G_{t2s})=e^{-\nabla x_{t}}||\nabla \tilde{y}_{tt}|| + e^{-\nabla x_{t}}||\nabla \tilde{y}_{st}||\\
\end{aligned}
\end{equation}
where $\nabla$ is the first derivative along spatial directions. We only apply the smoothness loss to $X_t$ and $X_{t2s}$ (real data), since $X_s$ and $X_{s2t}$ (synthetic data) have full supervision.\\
\noindent {\bf Depth Consistency Loss} We find that the predictions for $x_t$, {\it i.e.}, $F_t(x_t)$ and $F_s(G_{t2s}(x_t))$, show inconsistency in many regions, which is in contrast to our intuition. One of the possible reason is that $G_{t2s}$ might fail to translate $x_t$ with details. To enforce such coherence, we introduce an $\ell_1$ depth consistency loss with respect to $\tilde{y}_{tt}$ and $\tilde{y}_{st}$ as follows:
\begin{equation}
\begin{aligned}
\mathcal{L}_{dc}(F_t, F_s, G_{t2s})=||\tilde{y}_{tt}-\tilde{y}_{st}||.\\
\end{aligned}
\end{equation}

\noindent {\bf Full Objective} Our final loss function has the form as:
\begin{equation}
\begin{aligned}
\mathcal{L}&(G_{s2t}, G_{t2s}, D_t, D_s, F_t, F_s)\\
&=\mathcal{L}_{trans}(G_{s2t}, G_{t2s}, D_t, D_s)+\gamma_1 \mathcal{L}_{de}(F_t, F_s, G_{s2t})\\
&+\gamma_2\mathcal{L}_{gc}(F_t, F_s, G_{t2s})+\gamma_3 \mathcal{L}_{dc}(F_t, F_s, G_{t2s})\\
&+\gamma_4 \mathcal{L}_{ds}(F_t, F_s, G_{t2s})
\end{aligned}
\end{equation}
where $\gamma_n$($n\in\{1,2,3,4\}$) are trade-off factors. We optimize this objective function in an end-to-end deep network.
\subsection{Inference}
\label{sec:inference}
\begin{figure}[t]
\begin{center}
\includegraphics[width=1.0\linewidth]{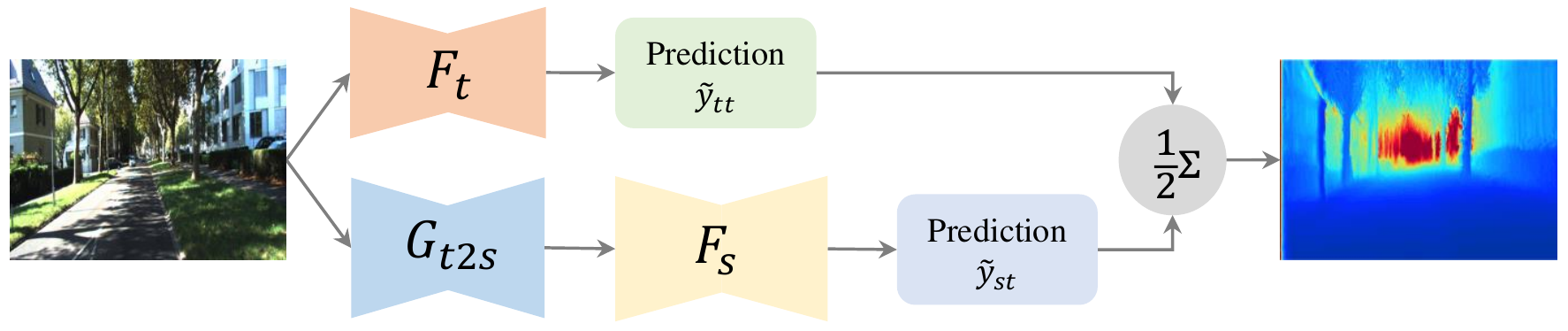}
\end{center}
\setlength{\abovecaptionskip}{-0.01pt}
\captionsetup{font={small}}
   \caption{Inference Phase (Section~\ref{sec:inference}).}
\label{fig:inference}
\end{figure}

In the inference phase, we aim to predict the depth map for a given image in real domain ({\it e.g.} KITTI dataset~\cite{menze2015object}) using the resultant models. In fact, there are two paths acquiring predicted depth maps: $x_t\to F_t(x_t)\to \tilde{y}_{tt}$ and $x_t\to G_{t2s}(x_t)\to x_{t2s}\to F_s(x_{t2s})\to \tilde{y}_{st}$, as shown in Figure~\ref{fig:inference}, and the final prediction is the average of $\tilde{y}_{tt}$ and $\tilde{y}_{st}$:
\begin{equation}
\begin{aligned}
\tilde{y}_t=\frac{1}{2}(\tilde{y}_{tt}+\tilde{y}_{st}).\\
\end{aligned}
\end{equation}
\section{Experiments}
In this section, we first present the details about our network architecture and the learning strategy. Then, we perform GASDA on one of the largest dataset in the context of autonomous driving, {\it i.e.}, KITTI dataset~\cite{menze2015object}. We also demonstrate the generalization capabilities of our model to other real-world scenes contained in Make3D~\cite{saxena2009make3d}. Finally, we conduct various ablations to analyze GASDA.
\begin{table*}[htp]\scriptsize
\centering
\begin{tabular}{c||c|c|c||cccc|ccc}
\hline
\multirow{2}{*}{Method} & \multirow{2}{*}{Supervised} & \multirow{2}{*}{Dataset} & \multirow{2}{*}{Cap} & \multicolumn{4}{c|}{Error Metrics (lower, better)} & \multicolumn{3}{c}{Accuracy Metrics (higher, better)} \\ \cline{5-11}
                  &                   &                   &                   &  Abs Rel   &  Sq Rel   &  RMSE   & RMSE log   &   $\delta<1.25$    &   $\delta<1.25^2$    &  $\delta<1.25^3$    \\
\hline\hline
  Eigen {\it et al.}~\cite{eigen2014depth}                &     Yes              &        K           &     $80m$              &  0.203   &  1.548   &  6.307   &  0.282  &   0.702    &   0.890    &  0.958    \\
  Liu {\it et al.}~\cite{liu2016learning}                &      Yes             &      K             &  $80m$                  &  0.202   &  1.614   &  6.523   & 0.275   &  0.678     &  0.895     &  0.965    \\
     Zhou {\it et al.}~\cite{zhou2017unsupervised}             &    No               &    K            &       $80m$            & 0.208    &  1.768   &  6.856   & 0.283   &  0.678     &   0.885    &  0.957    \\
     Zhou {\it et al.}~\cite{zhou2017unsupervised}              &   No                &    K+CS               &     $80m$              & 0.198    &  1.836   &  6.565   &  0.275  &   0.718    &   0.901    &  0.960    \\
     Kuznietsov {\it et al.}~\cite{kuznietsov2017semi}             &     Semi              &     K              &        $80m$           &  0.113   &  0.741   &  4.621   &  0.189  &  0.862     &   0.960    & 0.986     \\
     Godard {\it et al.}~\cite{godard2017unsupervised}             &     No              &     K              &     $80m$              &  0.148   &  1.344   &  5.927   & 0.247   &  0.803     &  0.922     & 0.964     \\
\hline
 All synthetic(baseline1)      &   No             &  S            &   $80m$             &  0.253   &  2.303   &   6.953   & 0.328   & 0.635    &  0.856    & 0.937 \\
All real(baseline2)&  No             &  K     &   $80m$             &  0.158   &  1.151   & 5.285   &  0.238   & 0.811    & 0.934    &  0.970\\
\hline
   \cellcolor[HTML]{EFEFEF} Kundu {\it et al.}~\cite{kundu2018adadepth}              &     \cellcolor[HTML]{EFEFEF} No             &     \cellcolor[HTML]{EFEFEF}  K+S(DA)            &   \cellcolor[HTML]{EFEFEF}   $80m$             & \cellcolor[HTML]{EFEFEF} 0.214   & \cellcolor[HTML]{EFEFEF} 1.932   & \cellcolor[HTML]{EFEFEF} 7.157   &\cellcolor[HTML]{EFEFEF}0.295   & \cellcolor[HTML]{EFEFEF}  0.665    & \cellcolor[HTML]{EFEFEF}  0.882    & \cellcolor[HTML]{EFEFEF} 0.950 \\
  \cellcolor[HTML]{EFEFEF}  Kundu {\it et al.}~\cite{kundu2018adadepth}              &  \cellcolor[HTML]{EFEFEF}    Semi             &   \cellcolor[HTML]{EFEFEF}    K+S(DA)            &  \cellcolor[HTML]{EFEFEF}    $80m$             & \cellcolor[HTML]{EFEFEF} 0.167   & \cellcolor[HTML]{EFEFEF} 1.257   &\cellcolor[HTML]{EFEFEF}  5.578   &\cellcolor[HTML]{EFEFEF} 0.237   & \cellcolor[HTML]{EFEFEF}  0.771    & \cellcolor[HTML]{EFEFEF}  0.922    & \cellcolor[HTML]{EFEFEF} 0.971 \\
   \cellcolor[HTML]{EFEFEF}  GASDA      &   \cellcolor[HTML]{EFEFEF}   No             & \cellcolor[HTML]{EFEFEF}      K+S(DA)           &   \cellcolor[HTML]{EFEFEF}   $80m$             & \cellcolor[HTML]{EFEFEF} {\bf 0.149}   & \cellcolor[HTML]{EFEFEF} {\bf 1.003}   & \cellcolor[HTML]{EFEFEF} {\bf 4.995}   &\cellcolor[HTML]{EFEFEF} {\bf 0.227}   &   \cellcolor[HTML]{EFEFEF}   {\bf 0.824}    & \cellcolor[HTML]{EFEFEF}  {\bf 0.941}    &\cellcolor[HTML]{EFEFEF} {\bf 0.973} \\
\hline \hline
Kuznietsov {\it et al.}~\cite{kuznietsov2017semi}             &     Yes              &     K              &        $50m$           &  0.117   &  0.597   &  3.531   &  0.183  &  0.861     &   0.964    & 0.989      \\
     Garg {\it et al.}~\cite{garg2016unsupervised} & No & K & $50m$ & 0.169 & 1.080 & 5.104 & 0.273 & 0.740 & 0.904 & 0.962 \\
     Godard {\it et al.}~\cite{godard2017unsupervised}             &     No              &     K              &     $50m$              &  0.140   &  0.976   &  4.471   & 0.232   &  0.818     &  0.931     & 0.969     \\
\hline
All synthetic(baseline1)      & No      &      S       &   $50m$             &  0.244   &  1.771   &  5.354   &  0.313   &   0.647    &  0.866    & 0.943 \\
     All real(baseline2)&    No       &     K           &    $50m$             & 0.151   &  0.856   &  4.043   & 0.227   & 0.824    &  0.940    &  0.973 \\
\hline
  \cellcolor[HTML]{EFEFEF}  Kundu {\it et al.}~\cite{kundu2018adadepth}              & \cellcolor[HTML]{EFEFEF}     No             &    \cellcolor[HTML]{EFEFEF}   K+S(DA)            &  \cellcolor[HTML]{EFEFEF}    $50m$             & \cellcolor[HTML]{EFEFEF} 0.203   & \cellcolor[HTML]{EFEFEF} 1.734   & \cellcolor[HTML]{EFEFEF} 6.251   & \cellcolor[HTML]{EFEFEF} 0.284   & \cellcolor[HTML]{EFEFEF}  0.687    & \cellcolor[HTML]{EFEFEF}  0.899    & \cellcolor[HTML]{EFEFEF} 0.958 \\
  \cellcolor[HTML]{EFEFEF}  Kundu {\it et al.}~\cite{kundu2018adadepth}              &   \cellcolor[HTML]{EFEFEF}   Semi             &   \cellcolor[HTML]{EFEFEF}    K+S(DA)            &   \cellcolor[HTML]{EFEFEF}   $50m$             & \cellcolor[HTML]{EFEFEF} 0.162   & \cellcolor[HTML]{EFEFEF} 1.041   & \cellcolor[HTML]{EFEFEF} 4.344   & \cellcolor[HTML]{EFEFEF} 0.225   & \cellcolor[HTML]{EFEFEF}  0.784    & \cellcolor[HTML]{EFEFEF}  0.930    & \cellcolor[HTML]{EFEFEF} 0.974 \\
  \cellcolor[HTML]{EFEFEF}    Zheng {\it et al.}~\cite{zheng2018t2net}              &  \cellcolor[HTML]{EFEFEF}    No             &  \cellcolor[HTML]{EFEFEF}     K+S(DA)            &  \cellcolor[HTML]{EFEFEF}    $50m$             & \cellcolor[HTML]{EFEFEF} 0.168   & \cellcolor[HTML]{EFEFEF} 1.199   & \cellcolor[HTML]{EFEFEF}  4.674   & \cellcolor[HTML]{EFEFEF} 0.243   &  \cellcolor[HTML]{EFEFEF} 0.772    &  \cellcolor[HTML]{EFEFEF} 0.912    &  \cellcolor[HTML]{EFEFEF} 0.966 \\
  \cellcolor[HTML]{EFEFEF}   GASDA      &   \cellcolor[HTML]{EFEFEF}   No             & \cellcolor[HTML]{EFEFEF}      K+S(DA)           & \cellcolor[HTML]{EFEFEF}     $50m$             & \cellcolor[HTML]{EFEFEF} {\bf 0.143}   & \cellcolor[HTML]{EFEFEF} {\bf 0.756}   & \cellcolor[HTML]{EFEFEF} {\bf 3.846}  & \cellcolor[HTML]{EFEFEF} {\bf 0.217}   &   \cellcolor[HTML]{EFEFEF} {\bf 0.836}    &  \cellcolor[HTML]{EFEFEF} {\bf 0.946}    & \cellcolor[HTML]{EFEFEF} {\bf 0.976} \\
\hline
\end{tabular}
\captionsetup{font={small}}
\caption{Results on KITTI dataset using the test split suggested in~\cite{eigen2014depth}. For the training data, K represents KITTI dataset, CS is CityScapes dataset~\cite{cordts2016cityscapes}, and S is vKITTI dataset. Methods, which apply domain adaptation techniques, are marked by the {\color{gray} gray}.}
\label{tb:eigen}
\end{table*}
\begin{figure*}
\centering
   \begin{subfigure}[b]{0.19\textwidth}
    \centering
     \includegraphics[width=0.98\linewidth,height=0.48in]{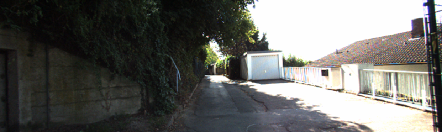}\vspace{1pt}\\
      \includegraphics[width=0.98\linewidth,height=0.48in]{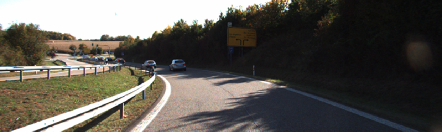}
      \subcaption*{Input Image}
    \end{subfigure}%
    \begin{subfigure}[b]{0.19\textwidth}
    \centering
     \includegraphics[width=0.98\linewidth,height=0.48in]{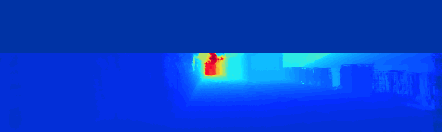}\vspace{1pt}\\
      \includegraphics[width=0.98\linewidth,height=0.48in]{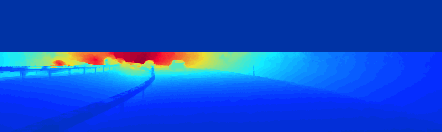}
      \subcaption*{Ground Truth}
    \end{subfigure}%
    \begin{subfigure}[b]{0.19\textwidth}
    \centering
     \includegraphics[width=0.98\linewidth,height=0.48in]{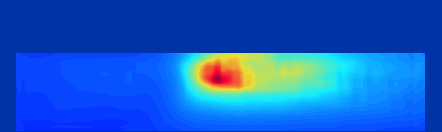}\vspace{1pt}\\
      \includegraphics[width=0.98\linewidth,height=0.48in]{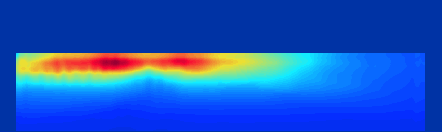}
      \subcaption*{Eigen {\it et.al.}~\cite{eigen2014depth}}
    \end{subfigure}
  \hspace{-5pt}
    \begin{subfigure}[b]{0.19\textwidth}
    \centering
      \includegraphics[width=0.98\linewidth,height=0.48in]{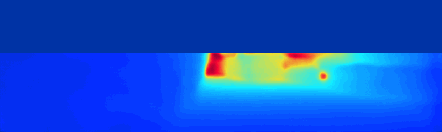}\vspace{1pt}\\
      \includegraphics[width=0.98\linewidth,height=0.48in]{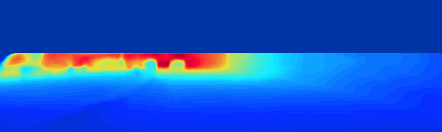}
      \subcaption*{Zheng {\it et.al.}~\cite{zheng2018t2net}}
    \end{subfigure}
    \hspace{-5.05pt}
    \begin{subfigure}[b]{0.19\textwidth}
    \centering
      \includegraphics[width=0.98\linewidth,height=0.48in]{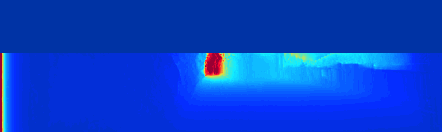}\vspace{1pt}\\
      \includegraphics[width=0.98\linewidth,height=0.48in]{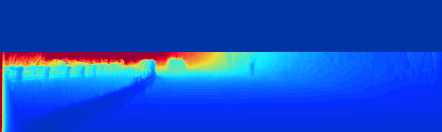}
      \subcaption*{GASDA}
    \end{subfigure}
  \captionsetup{font={small}}
  \caption{Qualitative comparison of our results against methods proposed by Eigen {\it et al.}~\cite{eigen2014depth} and Zheng {\it et al.}~\cite{zheng2018t2net} on KITTI. Ground truth has been interpolated for visualization. To facilitate comparison, we mask out the top regions, where ground truth depth is not available. Our approach preserves more details and yields high-quality depth maps.}
  \label{fig:example} 
\end{figure*}

\begin{table*}[htp]\small
\centering
\begin{tabular}{c||c|c||cccc|ccc}
\hline
\multirow{2}{*}{Method} & \multirow{2}{*}{Supervised} & \multirow{2}{*}{Dataset} & \multicolumn{4}{c|}{Error Metrics (lower, better)} & \multicolumn{3}{c}{Accuracy Metrics (higher, better)} \\ \cline{4-10}
                  &                   &                &  Abs Rel   &  Sq Rel   &  RMSE   & RMSE log   &   $\delta<1.25$    &   $\delta<1.25^2$    &  $\delta<1.25^3$    \\
\hline\hline
     Godard {\it et al.}~\cite{godard2017unsupervised}             &     No              &     K               &  0.124   &  1.388   &  6.125   &  0.217  &  0.841     &  0.936     & 0.975     \\
     Godard {\it et al.}~\cite{godard2017unsupervised}             &     No              &     K+CS             &  0.104   &  1.070   &  5.417   &  0.188  &  0.875     &  0.956     & 0.983     \\
     Atapour {\it et al.}~\cite{atapour2018real}            &      No             &       K+${\rm S^*}$(DA)    &  {\bf 0.101}   &  1.048   &  5.308   & 0.184   &   {\bf 0.903}    &   {\bf 0.988}    &  {\bf 0.992} \\
     GASDA &      No             &       K+S(DA)            &  0.106   &  {\bf 0.987}   &  {\bf 5.215}   & {\bf 0.176}   &   0.885    &   0.963    & 0.986 \\
\hline
\end{tabular}
\captionsetup{font={small}}
\caption{Results on $200$ training images of KITTI stereo 2015 benchmark~\cite{geiger2012we}. ${\rm S^*}$ is captured from GTA5, and more similar to real data than vKITTI. Our approach yields lower errors than state-of-the-art approaches, and achieve competitive accuracy compared with~\cite{atapour2018real}.}
\label{tb:kitti}
\end{table*}
\begin{figure}
\center
    \begin{subfigure}[b]{0.5\linewidth}
    \center
      \includegraphics[width=0.9\linewidth,height=0.8in]{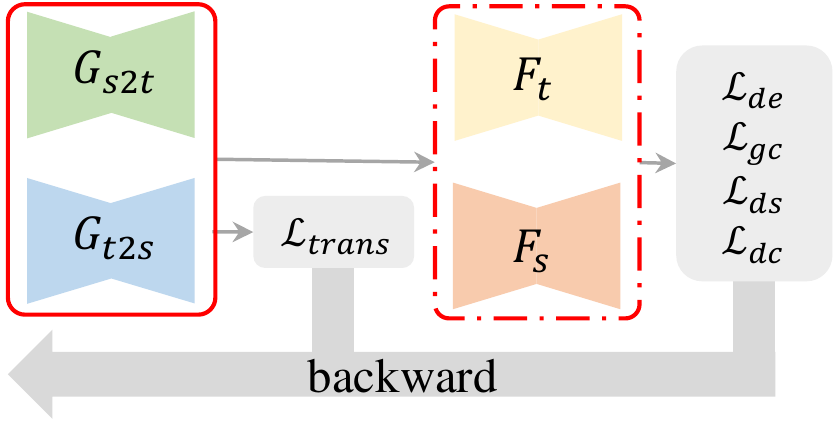}
      \subcaption*{Updating $G_{s2t}$ and $G_{t2s}$}
    \end{subfigure}%
    \vline
    \begin{subfigure}[b]{0.5\linewidth}
    \center
      \includegraphics[width=0.9\linewidth,height=0.8in]{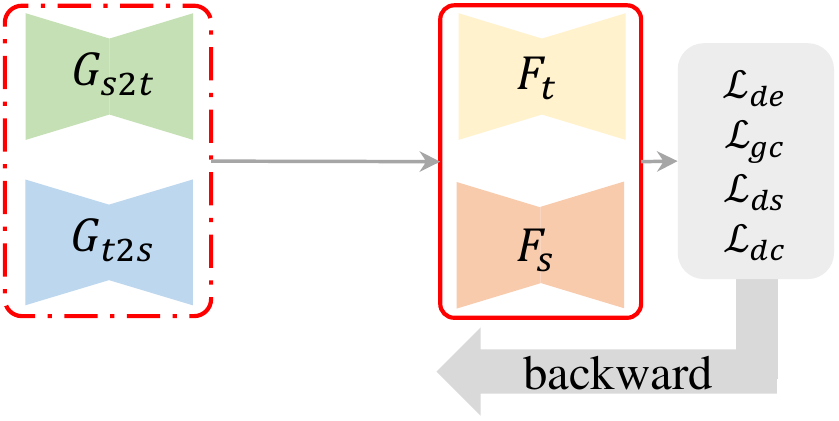}
      \subcaption*{Updating $F_t$ and $F_s$}
    \end{subfigure}
\captionsetup{font={small}}
  \caption{Iteratively updating stage. We learn our model by iteratively updating image style translators and depth estimators, {\it i.e.}, freezing the module with dashed box while updating the one with solidline box. See main text for details. We omit $D_t$ and $D_s$ for brevity.}
  \label{fig:training} 
\end{figure}
\begin{figure*}
\centering
    \begin{subfigure}[b]{0.16\textwidth}
    \centering
      \includegraphics[width=1.02\linewidth,height=0.41in]{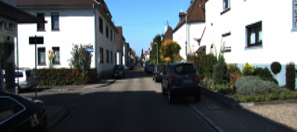}\vspace{1pt}\\
      \includegraphics[width=1.02\linewidth,height=0.41in]{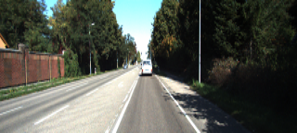}
      \subcaption*{Real Image}
    \end{subfigure}%
    \hspace{0.03pt}
    \begin{subfigure}[b]{0.16\textwidth}
    \centering
      \includegraphics[width=1.02\linewidth,height=0.41in]{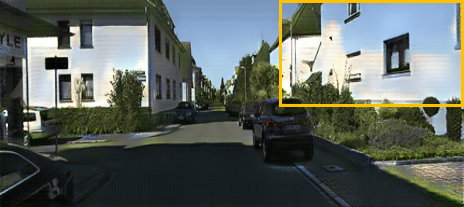}\vspace{1pt}\\
      \includegraphics[width=1.02\linewidth,height=0.41in]{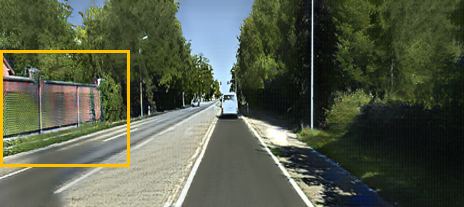}
      \subcaption*{CycleGAN~\cite{CycleGAN2017}}
    \end{subfigure}
    \begin{subfigure}[b]{0.16\textwidth}
    \centering
      \includegraphics[width=1.02\linewidth,height=0.41in]{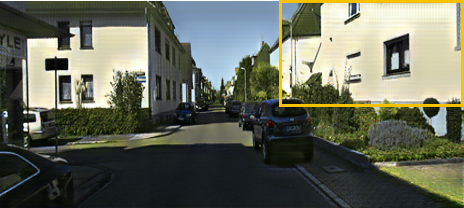}\vspace{1pt}\\
      \includegraphics[width=1.02\linewidth,height=0.41in]{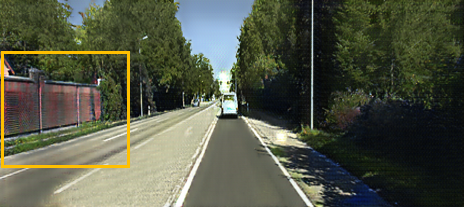}
      \subcaption*{GASDA}
    \end{subfigure}
  \hspace{-1.2pt}
  \vline
  \hspace{0.1pt}
    \begin{subfigure}[b]{0.16\textwidth}
    \centering
      \includegraphics[width=1.02\linewidth,height=0.41in]{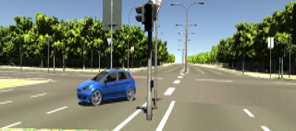}\vspace{1pt}\\
      \includegraphics[width=1.02\linewidth,height=0.41in]{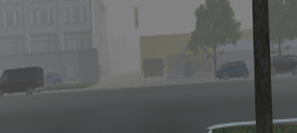}
      \subcaption*{Synthetic Image}
    \end{subfigure}
    \begin{subfigure}[b]{0.16\textwidth}
    \centering
      \includegraphics[width=1.02\linewidth,height=0.41in]{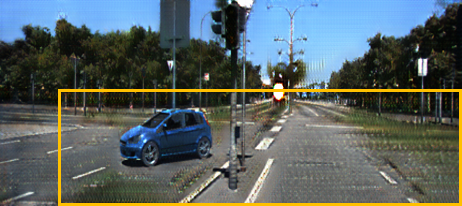}\vspace{1pt}\\
      \includegraphics[width=1.02\linewidth,height=0.41in]{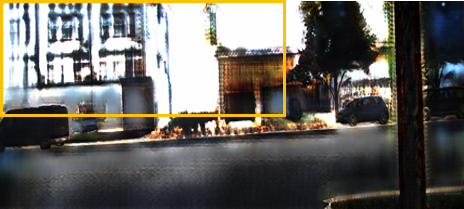}
      \subcaption*{CycleGAN~\cite{CycleGAN2017}}
    \end{subfigure}
    \begin{subfigure}[b]{0.16\textwidth}
    \centering
      \includegraphics[width=1.02\linewidth,height=0.41in]{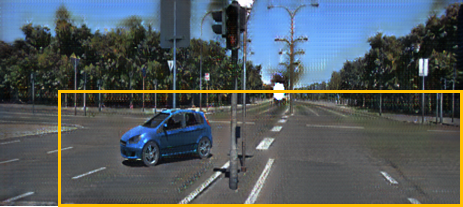}\vspace{1pt}\\
      \includegraphics[width=1.02\linewidth,height=0.41in]{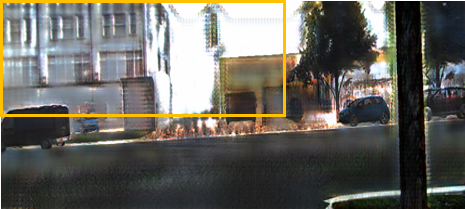}
      \subcaption*{GASDA}
    \end{subfigure}
  \captionsetup{font={small}}
  \caption{Qualitative image style translation results of our approach and CycleGAN~\cite{CycleGAN2017}. Left: real-to-synthetic translation; Right: synthetic-to-real translation. Our method can preserve geometric and semantic content better for both synthetic-to-real translation and the inverse one. Note that, the translation result is a by-product of GASDA. The improvement is marked by the yellow box.}
  \label{fig:trans} 
\end{figure*}
\begin{table*}[htp]\small
\centering
\begin{tabular}{c||cccc|ccc}
\hline
\multirow{2}{*}{Method} & \multicolumn{4}{c|}{Error Metrics (lower, better)} & \multicolumn{3}{c}{Accuracy Metrics (higher, better)} \\ \cline{2-8}
                                               &  Abs Rel   &  Sq Rel   &  RMSE   & RMSE log   &   $\delta<1.25$    & $\delta<1.25^2$    &  $\delta<1.25^3$    \\
\hline\hline
                       \multicolumn{8}{c}{\cellcolor[HTML]{EFEFEF} Domain Adaptation}\\
\hline
   SYN      & 0.253   & 2.303   & 6.953  & 0.328  & 0.635 & 0.856  &  0.937   \\
                                     SYN2REAL      & 0.229   & 2.094   & 6.530  & 0.294  & 0.691 & 0.886  &  0.951   \\
                                     SYN2REAL-E2E & {\bf 0.220}   & {\bf 1.969}   & {\bf 6.377}  & {\bf 0.284}  &  {\bf 0.703} & {\bf 0.895} & {\bf 0.956}  \\
\hline\hline
                       \multicolumn{8}{c}{\cellcolor[HTML]{EFEFEF} Geometry Consistency}\\
\hline

                                     REAL      & 0.158   & 1.151   & 5.285  & 0.238  & 0.811 & 0.934  &  0.970    \\
                                     SYN-GC    &     0.156 & 1.123  & 5.255 &  0.235  &  0.814  &  0.937  &   0.971  \\
                                     SYN2REAL-GC     &  0.153  & {\bf 1.112}   & {\bf 5.213}  & 0.233  & 0.819 & 0.938  & 0.972  \\
                                     SYN2REAL-GC-E2E     &  {\bf 0.152}  & 1.130   & 5.227  & {\bf 0.231}  & {\bf 0.821} & {\bf 0.939}  & {\bf 0.972}  \\
\hline\hline
                       \multicolumn{8}{c}{\cellcolor[HTML]{EFEFEF} Symmetric Domain Adaptation}    \\
\hline
                                     REAL2SYN-SYN-GC-E2E      &  0.160  & 1.226   & 5.412  & 0.240  & 0.806 & 0.933  &  0.969    \\
                                     GASDA-w/oDC     & 0.151   &  1.098  & 5.136  & 0.230  & 0.822 & 0.940  &  0.972 \\
                                    GASDA-$F_t$ & 0.150 & 1.014 & 5.041 & 0.228& {\bf 0.824} & {\bf 0.941}  &  {\bf 0.973} \\

                                     GASDA-$F_s$  &  0.156  &  1.087  & 5.157 & 0.235  & 0.813 & 0.936  & 0.971  \\
                                     GASDA     & {\bf 0.149}   &  {\bf 1.003}  & {\bf 4.995}  & {\bf 0.227}  & {\bf 0.824} & {\bf 0.941}  &  {\bf 0.973} \\
\hline
\end{tabular}
\captionsetup{font={small}}
\caption{Quantitative results for ablation study on KITTI dataset using the test split suggested in~\cite{eigen2014depth}. SYN, REAL, REAL2SYN, and SYN2REAL represent the model trained on $X_s$, $X_t$, $G_{t2s}(X_t)$, and $G_{s2t}(X_s)$; E2E represents the end-to-end training; GC and DC denote the geometry consistency and depth consistency, respectively; GASDA-$F_t$ ($F_s$) represents the output of $F_t$ ($F_s$) in GASDA.}
\label{tb:ablation}
\end{table*}
\begin{figure}
\center
    \begin{subfigure}[b]{0.3\linewidth}
    \centering
      \includegraphics[width=1.025\linewidth,height=0.50in]{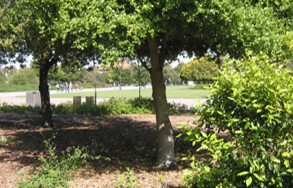}\vspace{1pt}\\
      \includegraphics[width=1.025\linewidth,height=0.50in]{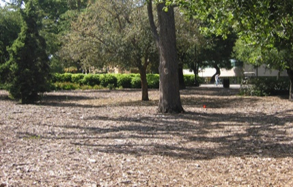}\vspace{1pt}\\
      \subcaption*{Input Image}
    \end{subfigure}
    \hspace{-0.01in}
    \begin{subfigure}[b]{0.3\linewidth}
    \centering
      \includegraphics[width=1.02\linewidth,height=0.50in]{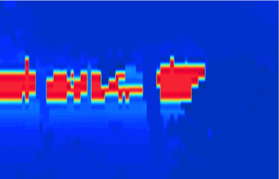}\vspace{1pt}\\
      \includegraphics[width=1.02\linewidth,height=0.50in]{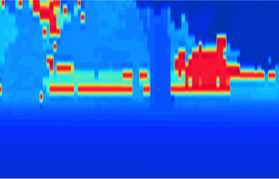}\vspace{1pt}\\
      \subcaption*{Ground Truth}
    \end{subfigure}
    \hspace{-0.021in}
    \begin{subfigure}[b]{0.3\linewidth}
    \centering
      \includegraphics[width=1.02\linewidth,height=0.50in]{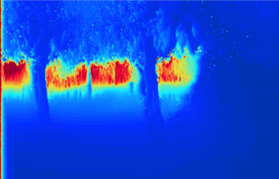}\vspace{1pt}\\
      \includegraphics[width=1.02\linewidth,height=0.50in]{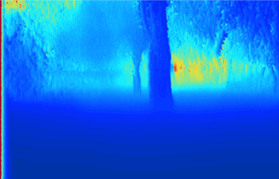}\vspace{1pt}\\
      \subcaption*{GASDA}
    \end{subfigure}
\captionsetup{font={small}}
  \caption{Qualitative results on Make3D dataset~\cite{saxena2009make3d}. Left to right: input image, ground truth depth, and our result.}
  \label{fig:make3d} 
\end{figure}
\subsection{Implementation Details}
\label{sec:ex}
\noindent {\bf Network Architecture} Our proposed framework consists of six sub-networks, which can be divided into three groups: $G_{s2t}$ and $G_{t2s}$ for image style translation, $D_t$ and $D_s$ for discrimination, $F_t$ and $F_s$ for monocular depth estimation. The networks in each group share the identical network architecture but are with different parameters. Specifically, we employ generators ($G_{s2t}$ and $G_{t2s}$) and discriminators ($D_s$ and $D_t$) provided by CycleGAN~\cite{CycleGAN2017}. For monocular depth estimators $F_t$ and $F_s$, we utilize the standard encoder-decoder structures with skip-connections and side outputs as~\cite{zheng2018t2net}.\\
\noindent {\bf Datasets} The target domain is KITTI~\cite{menze2015object}, which is a real-world computer vision benchmark consisting of $42,382$ rectified stereo pairs in the resolution about $375\times1242$. In our experiments,  the ground truth depth maps provided by KITTI are only for evaluation purpose. The source domain is Virtual KITTI (vKITTI)~\cite{gaidon2016virtual}, which contains $50$ photo-realistic synthetic videos with $21,260$ image-depth pairs of size $375\times1242$. Additionally, in order to study the generalization performance of our approach, we also apply the trained model to Make3D dataset~\cite{saxena2009make3d}. Since Make3D does not offer stereo images, we directly evaluate our model on the test split without training or further fine-tuning.

\noindent {\bf Training Details} We implement GASDA in {\it PyTorch}. We train our model in a two-stage manner, {\it i.e.}, a warming up stage and end-to-end iteratively updating stage.
In the warming up stage, we first optimize the style transfer networks for $10$ epochs with the momentum of $\beta_1 = 0.5$, $\beta_2 = 0.999$, and the initial learning rate of $\alpha = 0.0002$ using the ADAM solver~\cite{kingma2014adam}. Then we train $F_t$ on $\{X_t, G_{s2t}(X_s)\}$, and $F_s$ on $\{X_s, G_{t2s}(X_t)\}$ for around $20$ epochs by setting $\beta_1 = 0.9$, $\beta_2 = 0.999$, and $\alpha = 0.0001$. To make style translators generate high-quality images, so as to improve the subsequent depth estimators, we fine-tune the network in an end-to-end iteratively updating fashion as shown in Figure~\ref{fig:training}. In specific, we optimize $G_{s2t}$ and $G_{t2s}$ with the supervision of $F_t$ and $F_s$ for $m$ epochs, and then train $F_s$ and $F_t$ for $n$ epochs. We set $m=3$ and $n=7$ in our experiments, and repeat this process until the network converges (around $40$ epochs). In this stage, we employ the same momentum and solver as the first stage with the learning rates of $2e-6$ and $1e-5$ for the two respectively. The trade-off factors are set to $\lambda_1=10$, $\lambda_2=30$, $\gamma_1=50$, $\gamma_2=50$ and $\gamma_3=50$ and $\gamma_4=0.5$.
In the training phase, we down-sample all the images to $192\times640$, and increase the training set size using some common data augmentation strategies, including random horizontal flipping, rotation with the degrees of $[-5^\circ, 5^\circ]$, and brightness adjustment.
\subsection{KITTI Dataset}
We test our models on the $697$ images extracted from $29$ scenes, and  use all the $23,488$ images contained in other 32 scenes for training ($22,600$) and validation ($888$)~\cite{eigen2014depth,godard2017unsupervised}. To make a comparison with previous works, we evaluate our results in the regions with the ground truth depth less than $80m$ or $50m$ using standard error and accuracy metrics~\cite{godard2017unsupervised,zheng2018t2net}. Note that, the maximum depth value in vKITTI is $655.35m$ instead of $80m$ in KITTI, but unlike~\cite{zheng2018t2net}, we do not clip the depth maps of vKITTI to $80m$ during training.
\begin{table}\footnotesize
\centering
\begin{tabular}{c||c||ccc}
\hline
\multirow{2}{*}{Method} & \multirow{2}{*}{${\rm Trained^*}$} & \multicolumn{3}{c}{Error Metrics (lower, better)} \\ \cline{3-5}
                  &                           &  Abs Rel   &  Sq Rel   &  RMSE  \\
\hline\hline
     Karsch {\it et al.}~\cite{karsch2014depth}     &   Yes  & 0.398  & 4.723  & 7.801  \\
     Laina {\it et al.}~\cite{laina2016deeper}      &   Yes  & 0.198  & 1.665  & 5.461  \\
     Kundu {\it et al.}~\cite{kundu2018adadepth}    &   Yes  & 0.452  & 5.71   & 9.559  \\
\hline
     Godard {\it et al.}~\cite{godard2017unsupervised}   &   No  &   0.505   &  10.172   &  10.936    \\
     Kundu {\it et al.}~\cite{kundu2018adadepth}    &   No  & 0.647  & 12.341   & 11.567  \\
     Atapour {\it et al.}~\cite{atapour2018real}     &      No   &    0.423 &  9.343   &  {\bf 9.002}  \\
     GASDA &      No             &      {\bf 0.403}           &  {\bf 6.709}   &  10.424  \\
\hline
\end{tabular}
\captionsetup{font={small}}
\caption{Results on $134$ test images of Make3D~\cite{saxena2009make3d}. ${\rm Trained^*}$ indicates whether the model is trained on Make3D or not. Errors are computed for depths less than $70m$ in a central image crop~\cite{godard2017unsupervised}. It can be observed that our approach is comparable with those trained on Make3D.}
\label{tb:make3d}
\end{table}
In Table~\ref{tb:eigen},  we report the benchmark scores on the Eigen split~\cite{eigen2014depth} where the training sets are only KITTI and vKITTI. GASDA obtains a convincible improvement over previous state-of-the-art methods. Specifically, we make the comparisons with two baselines, {\it i.e.,} All synthetic (baseline1, trained on labeled synthetic data) and All real (baseline2, trained on real stereo pairs), and the latest domain adaptation methods~\cite{zheng2018t2net,kundu2018adadepth} and (semi-)supervised/unsupervised methods~\cite{eigen2014depth,liu2016learning,kuznietsov2017semi,garg2016unsupervised,godard2017unsupervised,zhou2017unsupervised}. The significant improvements in all the metrics demonstrate the superiority of our method. Note that, GASDA yields higher scores than~\cite{kundu2018adadepth} which employs additional ground truth depth maps for natural images contained in KITTI. GASDA cannot outperform~\cite{atapour2018real} in the Eigen split. The main reason is that the synthetic images employed in~\cite{atapour2018real} are captured from GTA5 \footnote{https://github.com/aitorzip/DeepGTAV.
}, and the domain shift between GTA5 and KITTI is not that significant than the one between vKITTI and KITTI. In addition, the training set size in~\cite{atapour2018real} is about three times than ours. However, GASDA performs competitively on the official KITTI stereo 2015 dataset and Make3D compared with~\cite{atapour2018real}, as reported in Table~\ref{tb:kitti} and Table~\ref{tb:make3d}. Apart from quantitative results, we also show some example outputs in Figure~\ref{fig:example}. Our approach preserves more details, and is able to recover depth information of small objects, such as the distant cars and rails, and generate clear boundaries.
\subsection{Make3D Dataset}
To discuss the generalization capabilities of GASDA, we evaluate our approach on Make3D dataset~\cite{saxena2009make3d} quantitatively and qualitatively. We do not train or further fine-tune our model using the images provide by Make3D. As shown in Table~\ref{tb:make3d} and Figure~\ref{fig:make3d}, although the domain shift between Make3D and KITTI is large, our model still performs well. Compared with state-of-the-art models~\cite{kundu2018adadepth,karsch2014depth,laina2016deeper} trained on Make3D in a supervised manner and others using domain adaptation~\cite{kundu2018adadepth,atapour2018real}, GASDA obtains impressive performance.
\subsection{Ablation Study}
Here, we conduct a series of ablations to analyze our approach. Quantitative results are shown in Table~\ref{tb:ablation}, and some sampled results for style transfer are shown in Figure~\ref{fig:trans}.

\noindent {\bf Domain Adaptation}
 We first demonstrate the effectiveness of domain adaptation by comparing two simple models, {\it i.e.} SYN ($F_s$ trained on $X_s$) and SYN2REAL ($F_t$ trained on $G_{s2t}(X_s)$). As shown in Table~\ref{tb:ablation}, SYN cannot capture satisfied scores on KITTI due to the domain shift. After the translation, the domain shift is reduced which means that the synthetic data distribution is relative closer to real data distribution. Thus, SYN2REAL is able to generalize better to real images. Further, we train the style translators ($G_{s2t}$ and $G_{t2s}$) and the depth estimation network ($F_t$) in an end-to-end fashion (SYN2REAL-E2E), which guides to a further improvement as compared to SYN2REAL. As a conclusion, the depth estimation network can improve the style transfer by providing a pixel-wise semantic constraint to the translation networks.
 Moreover, we can also observe the improvement in Figure~\ref{fig:trans} by comparing the translation results of original CycleGAN~\cite{CycleGAN2017} with ours.\\
\noindent {\bf Geometry Consistency} We then study the significance of the geometric constraint coming from stereo images based on the epipolar geometry. In specific, we employ the stereo images provided by KITTI when optimizing $F_t$ in SYN2REAL-E2E. We enforce the geometry consistency between the stereo images as a constraint as stated in Eq.~\ref{eq:gc}. The model SYN2REAL-GC-E2E outperforms SYN2REAL-E2E by a large margin, which demonstrates that the geometry consistency constraint can significantly improve standard domain adaptation frameworks. On the other hand,
the comparisons among SYN2REAL-GC, SYN-GC (trained on real data and synthetic data without domain adaptation) and REAL ($F_t$ trained on real stereo images without extra data) can show the significance of synthetic data with ground truth depth and domain adaptation.\\
\noindent {\bf Symmetric Domain Adaptation}  In contrast to previous works, we expect to fully take advantage of the bidirectional style translators $G_{s2t}$ and $G_{t2s}$. Thus, we learn REAL2SYN-SYN-GC-E2E whose network architecture is symmetrical to the aforementioned SYN2REAL-GC-E2E. We jointly optimized the two coupled with a depth consistency loss. As shown in Table~\ref{tb:ablation}, GASDA is superior than GASDA-w/oDC which demonstrates the effectiveness of the depth consistency loss. In addition, the comparisons (GASDA-$F_t$ {\it v.s.} SYN2ERAL-GC-E2E and GASDA-$F_s$ {\it v.s.} REAL2SYN-GC-E2E) show that the two can benefit each other in the jointly training.
\section{Conclusion}
In this paper, we present an unsupervised monocular depth estimation framework GASDA, which trains the monocular depth estimation model using the labelled synthetic data coupled with the epipolar geometry of real stereo data in a unified and symmetric deep learning network. Our main motivation is learning a depth estimation model from synthetic image-depth pairs in a supervised fashion, and at the same time taking into account the specific scene geometry information of the target data. Moreover, to alleviate the issues caused by domain shift, we reduce the domain discrepancy using the bidirectional image style transfer. Finally, we implement image translation and depth estimation in an end-to-end network so that then can improve each other. Experiments on KITTI and Make3D datasets show GASDA is able to generate desirable results quantitatively and qualitatively, and generalize well to unseen datasets.
\section{Acknowledgement}

This research was supported by Australian Research
Council Projects FL-170100117, DP-180103424 and IH-180100002.

{\small
\bibliographystyle{ieee}

\begin{thebibliography}{10}\itemsep=-1pt

\bibitem{ajakan2014domain}
Hana Ajakan, Pascal Germain, Hugo Larochelle, Fran{\c{c}}ois Laviolette, and
  Mario Marchand.
\newblock Domain-adversarial neural networks.
\newblock {\em arXiv preprint arXiv:1412.4446}, 2014.

\bibitem{atapour2018real}
Amir Atapour-Abarghouei and Toby~P Breckon.
\newblock Real-time monocular depth estimation using synthetic data with domain
  adaptation via image style transfer.
\newblock In {\em Proceedings of the IEEE Conference on Computer Vision and
  Pattern Recognition}, volume~18, page~1, 2018.

\bibitem{cao2016estimating}
Yuanzhouhan Cao, Zifeng Wu, and Chunhua Shen.
\newblock Estimating depth from monocular images as classification using deep
  fully convolutional residual networks.
\newblock {\em arXiv preprint arXiv:1605.02305}, 2016.

\bibitem{chen2016single}
Weifeng Chen, Zhao Fu, Dawei Yang, and Jia Deng.
\newblock Single-image depth perception in the wild.
\newblock In {\em Advances in Neural Information Processing Systems}, pages
  730--738, 2016.

\bibitem{chen2018domain}
Yuhua Chen, Wen Li, Christos Sakaridis, Dengxin Dai, and Luc Van~Gool.
\newblock Domain adaptive faster r-cnn for object detection in the wild.
\newblock In {\em Proceedings of the IEEE Conference on Computer Vision and
  Pattern Recognition}, pages 3339--3348, 2018.

\bibitem{cordts2016cityscapes}
Marius Cordts, Mohamed Omran, Sebastian Ramos, Timo Rehfeld, Markus Enzweiler,
  Rodrigo Benenson, Uwe Franke, Stefan Roth, and Bernt Schiele.
\newblock The cityscapes dataset for semantic urban scene understanding.
\newblock In {\em Proceedings of the IEEE conference on computer vision and
  pattern recognition}, pages 3213--3223, 2016.

\bibitem{dosovitskiy2015flownet}
Alexey Dosovitskiy, Philipp Fischer, Eddy Ilg, Philip Hausser, Caner Hazirbas,
  Vladimir Golkov, Patrick Van Der~Smagt, Daniel Cremers, and Thomas Brox.
\newblock Flownet: Learning optical flow with convolutional networks.
\newblock In {\em Proceedings of the IEEE International Conference on Computer
  Vision}, pages 2758--2766, 2015.

\bibitem{eigen2015predicting}
David Eigen and Rob Fergus.
\newblock Predicting depth, surface normals and semantic labels with a common
  multi-scale convolutional architecture.
\newblock In {\em Proceedings of the IEEE International Conference on Computer
  Vision}, pages 2650--2658, 2015.

\bibitem{eigen2014depth}
David Eigen, Christian Puhrsch, and Rob Fergus.
\newblock Depth map prediction from a single image using a multi-scale deep
  network.
\newblock In {\em Advances in neural information processing systems}, pages
  2366--2374, 2014.

\bibitem{fu2018deep}
Huan Fu, Mingming Gong, Chaohui Wang, Kayhan Batmanghelich, and Dacheng Tao.
\newblock Deep ordinal regression network for monocular depth estimation.
\newblock In {\em Proceedings of the IEEE Conference on Computer Vision and
  Pattern Recognition}, pages 2002--2011, 2018.

\bibitem{gaidon2016virtual}
Adrien Gaidon, Qiao Wang, Yohann Cabon, and Eleonora Vig.
\newblock Virtual worlds as proxy for multi-object tracking analysis.
\newblock In {\em Proceedings of the IEEE conference on computer vision and
  pattern recognition}, pages 4340--4349, 2016.

\bibitem{ganin2015unsupervised}
Yaroslav Ganin and Victor~S. Lempitsky.
\newblock Unsupervised domain adaptation by backpropagation.
\newblock In {\em ICML}, 2015.

\bibitem{ganin2016domain}
Yaroslav Ganin, Evgeniya Ustinova, Hana Ajakan, Pascal Germain, Hugo
  Larochelle, Fran{\c{c}}ois Laviolette, Mario Marchand, and Victor Lempitsky.
\newblock Domain-adversarial training of neural networks.
\newblock {\em The Journal of Machine Learning Research}, 17(1):2096--2030,
  2016.

\bibitem{garg2016unsupervised}
Ravi Garg, Vijay~Kumar BG, Gustavo Carneiro, and Ian Reid.
\newblock Unsupervised cnn for single view depth estimation: Geometry to the
  rescue.
\newblock In {\em European Conference on Computer Vision}, pages 740--756.
  Springer, 2016.

\bibitem{geiger2012we}
Andreas Geiger, Philip Lenz, and Raquel Urtasun.
\newblock Are we ready for autonomous driving? the kitti vision benchmark
  suite.
\newblock In {\em Computer Vision and Pattern Recognition (CVPR), 2012 IEEE
  Conference on}, pages 3354--3361. IEEE, 2012.

\bibitem{godard2017unsupervised}
Cl{\'e}ment Godard, Oisin Mac~Aodha, and Gabriel~J Brostow.
\newblock Unsupervised monocular depth estimation with left-right consistency.
\newblock In {\em CVPR}, volume~2, page~7, 2017.

\bibitem{gong2012geodesic}
Boqing Gong, Yuan Shi, Fei Sha, and Kristen Grauman.
\newblock Geodesic flow kernel for unsupervised domain adaptation.
\newblock In {\em Computer Vision and Pattern Recognition (CVPR), 2012 IEEE
  Conference on}, pages 2066--2073. IEEE, 2012.

\bibitem{gong2018causal}
Mingming Gong, Kun Zhang, Biwei Huang, Clark Glymour, Dacheng Tao, and Kayhan
  Batmanghelich.
\newblock Causal generative domain adaptation networks.
\newblock {\em arXiv preprint arXiv:1804.04333}, 2018.

\bibitem{gong2016domain}
Mingming Gong, Kun Zhang, Tongliang Liu, Dacheng Tao, Clark Glymour, and
  Bernhard Sch{\"o}lkopf.
\newblock Domain adaptation with conditional transferable components.
\newblock In {\em International conference on machine learning}, pages
  2839--2848, 2016.

\bibitem{goodfellow2014generative}
Ian Goodfellow, Jean Pouget-Abadie, Mehdi Mirza, Bing Xu, David Warde-Farley,
  Sherjil Ozair, Aaron Courville, and Yoshua Bengio.
\newblock Generative adversarial nets.
\newblock In {\em Advances in neural information processing systems}, pages
  2672--2680, 2014.

\bibitem{gretton2012kernel}
Arthur Gretton, Karsten~M Borgwardt, Malte~J Rasch, Bernhard Sch{\"o}lkopf, and
  Alexander Smola.
\newblock A kernel two-sample test.
\newblock {\em Journal of Machine Learning Research}, 13(Mar):723--773, 2012.

\bibitem{he2018learning}
Lei He, Guanghui Wang, and Zhanyi Hu.
\newblock Learning depth from single images with deep neural network embedding
  focal length.
\newblock {\em IEEE Transactions on Image Processing}, 2018.

\bibitem{jaderberg2015spatial}
Max Jaderberg, Karen Simonyan, Andrew Zisserman, et~al.
\newblock Spatial transformer networks.
\newblock In {\em Advances in neural information processing systems}, pages
  2017--2025, 2015.

\bibitem{karsch2014depth}
Kevin Karsch, Ce Liu, and Sing~Bing Kang.
\newblock Depth transfer: Depth extraction from video using non-parametric
  sampling.
\newblock {\em IEEE transactions on pattern analysis and machine intelligence},
  36(11):2144--2158, 2014.

\bibitem{kingma2014adam}
Diederik~P Kingma and Jimmy~Lei Ba.
\newblock Adam: Amethod for stochastic optimization.
\newblock In {\em Proc. 3rd Int. Conf. Learn. Representations}, 2014.

\bibitem{kundu2018adadepth}
Jogendra~Nath Kundu, Phani~Krishna Uppala, Anuj Pahuja, and R~Venkatesh Babu.
\newblock Adadepth: Unsupervised content congruent adaptation for depth
  estimation.
\newblock {\em arXiv preprint arXiv:1803.01599}, 2018.

\bibitem{kuznietsov2017semi}
Yevhen Kuznietsov, J{\"o}rg St{\"u}ckler, and Bastian Leibe.
\newblock Semi-supervised deep learning for monocular depth map prediction.
\newblock In {\em Proc. of the IEEE Conference on Computer Vision and Pattern
  Recognition}, pages 6647--6655, 2017.

\bibitem{ladicky2014pulling}
Lubor Ladicky, Jianbo Shi, and Marc Pollefeys.
\newblock Pulling things out of perspective.
\newblock In {\em Proceedings of the IEEE Conference on Computer Vision and
  Pattern Recognition}, pages 89--96, 2014.

\bibitem{lai2017semi}
Wei-Sheng Lai, Jia-Bin Huang, and Ming-Hsuan Yang.
\newblock Semi-supervised learning for optical flow with generative adversarial
  networks.
\newblock In {\em Advances in Neural Information Processing Systems}, pages
  354--364, 2017.

\bibitem{laina2016deeper}
Iro Laina, Christian Rupprecht, Vasileios Belagiannis, Federico Tombari, and
  Nassir Navab.
\newblock Deeper depth prediction with fully convolutional residual networks.
\newblock In {\em 3D Vision (3DV), 2016 Fourth International Conference on},
  pages 239--248. IEEE, 2016.

\bibitem{li2015depth}
Bo Li, Chunhua Shen, Yuchao Dai, Anton Van Den~Hengel, and Mingyi He.
\newblock Depth and surface normal estimation from monocular images using
  regression on deep features and hierarchical crfs.
\newblock In {\em Proceedings of the IEEE Conference on Computer Vision and
  Pattern Recognition}, pages 1119--1127, 2015.

\bibitem{li2018deep}
Ya Li, Xinmei Tian, Mingming Gong, Yajing Liu, Tongliang Liu, Kun Zhang, and
  Dacheng Tao.
\newblock Deep domain generalization via conditional invariant adversarial
  networks.
\newblock In {\em Proceedings of the European Conference on Computer Vision
  (ECCV)}, pages 624--639, 2018.

\bibitem{liu2010single}
Beyang Liu, Stephen Gould, and Daphne Koller.
\newblock Single image depth estimation from predicted semantic labels.
\newblock In {\em Computer Vision and Pattern Recognition (CVPR), 2010 IEEE
  Conference on}, pages 1253--1260. IEEE, 2010.

\bibitem{liu2011sift}
Ce Liu, Jenny Yuen, and Antonio Torralba.
\newblock Sift flow: Dense correspondence across scenes and its applications.
\newblock {\em IEEE transactions on pattern analysis and machine intelligence},
  33(5):978--994, 2011.

\bibitem{liu2016learning}
Fayao Liu, Chunhua Shen, Guosheng Lin, and Ian~D Reid.
\newblock Learning depth from single monocular images using deep convolutional
  neural fields.
\newblock {\em IEEE Trans. Pattern Anal. Mach. Intell.}, 38(10):2024--2039,
  2016.

\bibitem{liu2014discrete}
Miaomiao Liu, Mathieu Salzmann, and Xuming He.
\newblock Discrete-continuous depth estimation from a single image.
\newblock In {\em Proceedings of the IEEE Conference on Computer Vision and
  Pattern Recognition}, pages 716--723, 2014.

\bibitem{long2013transfer}
Mingsheng Long, Guiguang Ding, Jianmin Wang, Jiaguang Sun, Yuchen Guo, and
  Philip~S Yu.
\newblock Transfer sparse coding for robust image representation.
\newblock In {\em Proceedings of the IEEE conference on computer vision and
  pattern recognition}, pages 407--414, 2013.

\bibitem{menze2015object}
Moritz Menze and Andreas Geiger.
\newblock Object scene flow for autonomous vehicles.
\newblock In {\em Proceedings of the IEEE Conference on Computer Vision and
  Pattern Recognition}, pages 3061--3070, 2015.

\bibitem{pan2010survey}
Sinno~Jialin Pan, Qiang Yang, et~al.
\newblock A survey on transfer learning.
\newblock {\em IEEE Transactions on knowledge and data engineering},
  22(10):1345--1359, 2010.

\bibitem{qi2018geonet}
Xiaojuan Qi, Renjie Liao, Zhengzhe Liu, Raquel Urtasun, and Jiaya Jia.
\newblock Geonet: Geometric neural network for joint depth and surface normal
  estimation.
\newblock In {\em Proceedings of the IEEE Conference on Computer Vision and
  Pattern Recognition}, pages 283--291, 2018.

\bibitem{repala2018dual}
Vamshi~Krishna Repala and Shiv~Ram Dubey.
\newblock Dual cnn models for unsupervised monocular depth estimation.
\newblock {\em arXiv preprint arXiv:1804.06324}, 2018.

\bibitem{roy2016monocular}
Anirban Roy and Sinisa Todorovic.
\newblock Monocular depth estimation using neural regression forest.
\newblock In {\em Proceedings of the IEEE Conference on Computer Vision and
  Pattern Recognition}, pages 5506--5514, 2016.

\bibitem{saenko2010adapting}
Kate Saenko, Brian Kulis, Mario Fritz, and Trevor Darrell.
\newblock Adapting visual category models to new domains.
\newblock In {\em European conference on computer vision}, pages 213--226.
  Springer, 2010.

\bibitem{saxena2006learning}
Ashutosh Saxena, Sung~H Chung, and Andrew~Y Ng.
\newblock Learning depth from single monocular images.
\newblock In {\em Advances in neural information processing systems}, pages
  1161--1168, 2006.

\bibitem{saxena2009make3d}
Ashutosh Saxena, Min Sun, and Andrew~Y Ng.
\newblock Make3d: Learning 3d scene structure from a single still image.
\newblock {\em IEEE transactions on pattern analysis and machine intelligence},
  31(5):824--840, 2009.

\bibitem{sun2016return}
Baochen Sun, Jiashi Feng, and Kate Saenko.
\newblock Return of frustratingly easy domain adaptation.
\newblock In {\em AAAI}, volume~6, page~8, 2016.

\bibitem{sun2016deep}
Baochen Sun and Kate Saenko.
\newblock Deep coral: Correlation alignment for deep domain adaptation.
\newblock In {\em European Conference on Computer Vision}, pages 443--450.
  Springer, 2016.

\bibitem{taigman2016unsupervised}
Yaniv Taigman, Adam Polyak, and Lior Wolf.
\newblock Unsupervised cross-domain image generation.
\newblock {\em arXiv preprint arXiv:1611.02200}, 2016.

\bibitem{torralba2011unbiased}
Antonio Torralba and Alexei~A Efros.
\newblock Unbiased look at dataset bias.
\newblock In {\em Computer Vision and Pattern Recognition (CVPR), 2011 IEEE
  Conference on}, pages 1521--1528. IEEE, 2011.

\bibitem{tzeng2014deep}
Eric Tzeng, Judy Hoffman, Ning Zhang, Kate Saenko, and Trevor Darrell.
\newblock Deep domain confusion: Maximizing for domain invariance.
\newblock {\em arXiv preprint arXiv:1412.3474}, 2014.

\bibitem{wang2015towards}
Peng Wang, Xiaohui Shen, Zhe Lin, Scott Cohen, Brian Price, and Alan~L Yuille.
\newblock Towards unified depth and semantic prediction from a single image.
\newblock In {\em Proceedings of the IEEE Conference on Computer Vision and
  Pattern Recognition}, pages 2800--2809, 2015.

\bibitem{wang2004image}
Zhou Wang, Alan~C Bovik, Hamid~R Sheikh, and Eero~P Simoncelli.
\newblock Image quality assessment: from error visibility to structural
  similarity.
\newblock {\em IEEE transactions on image processing}, 13(4):600--612, 2004.

\bibitem{xie2016deep3d}
Junyuan Xie, Ross Girshick, and Ali Farhadi.
\newblock Deep3d: Fully automatic 2d-to-3d video conversion with deep
  convolutional neural networks.
\newblock In {\em European Conference on Computer Vision}, pages 842--857.
  Springer, 2016.

\bibitem{xu2017multi}
Dan Xu, Elisa Ricci, Wanli Ouyang, Xiaogang Wang, and Nicu Sebe.
\newblock Multi-scale continuous crfs as sequential deep networks for monocular
  depth estimation.
\newblock In {\em Proceedings of CVPR}, volume~1, 2017.

\bibitem{xu2018structured}
Dan Xu, Wei Wang, Hao Tang, Hong Liu, Nicu Sebe, and Elisa Ricci.
\newblock Structured attention guided convolutional neural fields for monocular
  depth estimation.
\newblock In {\em Proceedings of the IEEE Conference on Computer Vision and
  Pattern Recognition}, pages 3917--3925, 2018.

\bibitem{yin2018geonet}
Zhichao Yin and Jianping Shi.
\newblock Geonet: Unsupervised learning of dense depth, optical flow and camera
  pose.
\newblock In {\em Proceedings of the IEEE Conference on Computer Vision and
  Pattern Recognition (CVPR)}, volume~2, 2018.

\bibitem{zhan2018unsupervised}
Huangying Zhan, Ravi Garg, Chamara~Saroj Weerasekera, Kejie Li, Harsh Agarwal,
  and Ian Reid.
\newblock Unsupervised learning of monocular depth estimation and visual
  odometry with deep feature reconstruction.
\newblock In {\em Proceedings of the IEEE Conference on Computer Vision and
  Pattern Recognition}, pages 340--349, 2018.

\bibitem{zhang2013domain}
Kun Zhang, Bernhard Sch{\"o}lkopf, Krikamol Muandet, and Zhikun Wang.
\newblock Domain adaptation under target and conditional shift.
\newblock In {\em International Conference on Machine Learning}, pages
  819--827, 2013.

\bibitem{zheng2018t2net}
Chuanxia Zheng, Tat-Jen Cham, and Jianfei Cai.
\newblock T2net: Synthetic-to-realistic translation for solving single-image
  depth estimation tasks.
\newblock In {\em Proceedings of the European Conference on Computer Vision
  (ECCV)}, pages 767--783, 2018.

\bibitem{zhou2017unsupervised}
Tinghui Zhou, Matthew Brown, Noah Snavely, and David~G Lowe.
\newblock Unsupervised learning of depth and ego-motion from video.
\newblock In {\em CVPR}, volume~2, page~7, 2017.

\bibitem{CycleGAN2017}
Jun-Yan Zhu, Taesung Park, Phillip Isola, and Alexei~A Efros.
\newblock Unpaired image-to-image translation using cycle-consistent
  adversarial networkss.
\newblock In {\em Computer Vision (ICCV), 2017 IEEE International Conference
  on}, 2017.

\end{thebibliography}

}
\end{document}